\documentclass[conference]{IEEEtran}
\IEEEoverridecommandlockouts

\usepackage{cite}
\usepackage{amsmath,amssymb,amsfonts}
\usepackage{algorithmic}
\usepackage{graphicx}
\usepackage{textcomp}
\usepackage{xcolor}
\usepackage{multirow}
\usepackage{booktabs}
\usepackage{stfloats}
\graphicspath{{./}}

\def\BibTeX{{\rm B\kern-.05em{\sc i\kern-.025em b}\kern-.08em
    T\kern-.1667em\lower.7ex\hbox{E}\kern-.125emX}}

\begin{document}

\title{CGF-DETR: Cross-Gated Fusion DETR for Enhanced Pneumonia Detection in Chest X-rays}

%

\author{
	Yefeng Wu\textsuperscript{1}\textsuperscript{,}\textsuperscript{*}, Yuchen Song\textsuperscript{2}, Ling Wu\textsuperscript{1}, Shan Wan\textsuperscript{1}, Yecheng Zhao\textsuperscript{1} \\
	\textsuperscript{1}Electronic Science and Technology, Anhui University, Hefei, China \\
	\textsuperscript{2}Medical Imaging Science, Wannan Medical College, Wuhu, China \\
	\textsuperscript{*}Emails: wuyefengflc@163.com
}

\maketitle

\begin{abstract}
Pneumonia remains a leading cause of morbidity and mortality worldwide, necessitating accurate and efficient automated detection systems. While recent transformer-based detectors like RT-DETR have shown promise in object detection tasks, their application to medical imaging, particularly pneumonia detection in chest X-rays, remains underexplored. This paper presents CGF-DETR, an enhanced real-time detection transformer specifically designed for pneumonia detection. We introduce XFABlock in the backbone to improve multi-scale feature extraction through convolutional attention mechanisms integrated with CSP architecture. To achieve efficient feature aggregation, we propose SPGA module that replaces standard multi-head attention with dynamic gating mechanisms and single-head self-attention. Additionally, GCFC3 is designed for the neck to enhance feature representation through multi-path convolution fusion while maintaining real-time performance via structural re-parameterization. Extensive experiments on the RSNA Pneumonia Detection dataset demonstrate that CGF-DETR achieves 82.2\% mAP@0.5, outperforming the baseline RT-DETR-l by 3.7\% while maintaining comparable inference speed at 48.1 FPS. Our ablation studies confirm that each proposed module contributes meaningfully to the overall performance improvement, with the complete model achieving 50.4\% mAP@[0.5:0.95].
\end{abstract}


\section{Introduction}

Pneumonia is an acute respiratory infection affecting the lungs, causing significant health burdens globally with millions of cases annually. Early and accurate diagnosis is critical for effective treatment and reducing mortality rates\cite{Kulkarni2023}.  Chest X-ray (CXR) imaging remains the primary diagnostic tool due to its accessibility and cost-effectiveness \cite{Sharma2024}. However, manual interpretation of CXRs is time-consuming, subject to inter-observer variability, and requires specialized expertise that may not be readily available in resource-limited settings.

Deep learning has revolutionized medical image analysis, with convolutional neural networks (CNNs) demonstrating radiologist-level performance in various diagnostic tasks \cite{Akhter2023,Iqbal2024}. Recent advances in object detection, particularly transformer-based approaches like DETR (DEtection TRansformer) \cite{Carion2020}, have shown remarkable success by formulating detection as a direct set prediction problem. RT-DETR further improves upon this by achieving real-time performance through efficient encoder designs and hybrid encoding strategies, making it particularly suitable for practical medical applications where both accuracy and speed are critical.

Despite these advances, applying general-purpose object detectors to medical imaging presents unique challenges \cite{Ali2024}. Pneumonia manifestations in chest X-rays often appear as subtle opacity patterns with varying shapes, sizes, and locations \cite{Dey2021,Mabrouk2022}. These lesions may be small, have low contrast against lung tissue, and can be easily confused with normal anatomical structures. Furthermore, medical imaging requires models that can capture both fine-grained local features and global contextual information effectively.

To address these challenges, we propose CGF-DETR (Cross-Gated Fusion DETR), a specialized architecture that enhances RT-DETR for pneumonia detection through three synergistic innovations: the Cross-Fuse Attention Block (XFABlock) for enhanced multi-scale feature extraction in the backbone, the Split-Path Gated Attention (SPGA) module for efficient feature aggregation in the encoder, and the Generalized Convolution Fusion RepC3 (GCFC3) for rich multi-scale representation in the neck.

Our main contributions are summarized as follows:

\begin{itemize}
\item We propose XFABlock, which integrates convolutional attention mechanisms with CSP architecture to enhance multi-scale feature extraction in the backbone. By employing XFABlock modules that combine spatial attention through depthwise convolutions with residual connections and FFN transformations, XFABlock enables the model to better capture subtle pneumonia patterns at multiple scales while maintaining efficient gradient flow.

\item We introduce SPGA, a novel split-path gated attention mechanism. SPGA employs single-head self-attention with dynamic gating to adaptively control attention sparsity based on input content, improving feature fusion. In our experiments, SPGA alone yields +1.1\% mAP@0.5 with a small overhead; combined with GCFC3, the overall model latency decreases relative to the baseline.

\item We design GCFC3, a generalized convolution fusion module that enhances multi-scale feature representation in the neck through multi-path convolution fusion. By maintaining multiple parallel convolution paths during training and fusing them into a single 3×3 convolution via structural re-parameterization at inference, GCFC3 achieves rich feature learning capacity without sacrificing deployment efficiency.
\end{itemize}

\section{Related Work}

\subsection{Medical Image Analysis for Pneumonia Detection}

Pneumonia detection from chest X-rays has progressed from handcrafted features with classical classifiers to deep CNNs that deliver strong performance (CheXNet and variants based on ResNet, DenseNet, EfficientNet). However, most works frame the task as image-level classification \cite{ElGhandour2024} and thus lack precise spatial evidence; weakly supervised CAM or attention methods provide only coarse heatmaps \cite{Selvaraju2017,WangHaofan2020}. Object detection frameworks directly localize lesions with bounding boxes, offering clinically useful interpretability and decision support, which motivates our detector-based formulation \cite{Xie2025}.

\begin{figure*}[!b]
\centering
\includegraphics[width=0.9\textwidth]{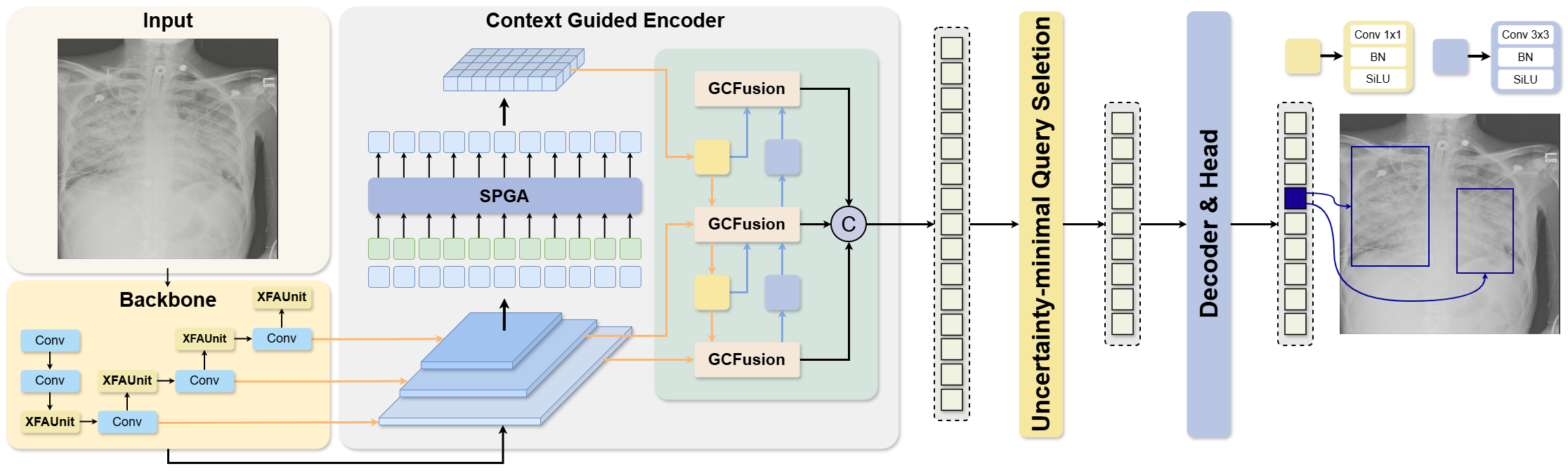}
\caption{Overall architecture of CGF-DETR. The backbone integrates XFABlock, the encoder leverages SPGA, the neck adopts GCFC3, and detection heads follow the RT-DETR design.}
\label{fig:architecture}
\end{figure*}

\subsection{Object Detection Methods}

Modern detectors evolved from sliding windows to two-stage R-CNN and Faster R-CNN\cite{Ren2015}, one-stage SSD\cite{Lin2017}, YOLO and RetinaNet\cite{WangLiao2024}, and anchor-free designs FCOS and CenterNet \cite{Tian2019,Duan2019}. Multi-scale representation is strengthened by FPN-like pyramids with CSP-style backbones. Training has benefited from focal-style classification, IoU-based box losses, and improved assignment \cite{Zaidi2022}. Post-processing typically relies on NMS and its variants, which are threshold-sensitive and can suppress true positives in crowded scenes \cite{Wu2024}. The YOLO family and RTMDet push the real-time frontier via decoupled heads, lightweight blocks, and stronger augmentation, yielding solid accuracy–latency trade-offs on natural images \cite{WangLiao2024,Liu2023,WangBochkovskiy2023}. Yet dense CNN heads still depend on anchors or priors and heuristic NMS, require dataset-specific tuning, and suffer class imbalance and duplicate predictions—motivating end-to-end transformer detectors that cut hand-crafted components and capture long-range dependencies.

\subsection{DETR Network}

DETR formulates detection as set prediction: CNN features go into a transformer encoder–decoder with learnable queries, and predictions are matched to ground truth via Hungarian assignment with set losses, removing anchors and NMS. Global attention on high-resolution features makes vanilla DETR slow to converge and weak on small objects. Follow-ups ease this: Deformable DETR adds multi-scale sparse attention around reference points; Conditional DETR refine query–key conditioning; DINO accelerates learning with denoising and mixed queries. Still, many transformer detectors are compute-heavy and often miss real-time targets. RT-DETR narrows the gap by decoupling intra-scale interaction from cross-scale fusion and using IoU-aware query selection, balancing accuracy and speed \cite{Zhao2024}. In medical imaging, subtle low-contrast patterns, large scale variation, and the need for efficient cross-scale aggregation under real-time limits remain open.

\section{Methodology}

\subsection{Overall Architecture}

CGF-DETR follows the RT-DETR architecture framework, consisting of a backbone for feature extraction, an encoder neck for multi-scale feature fusion, and a transformer-based decoder for object queries. Our key modifications are strategically placed to enhance feature extraction and fusion capabilities while maintaining real-time performance. Fig. \ref{fig:architecture} illustrates the overall architecture.

The backbone employs a hierarchical structure with four stages, progressively downsampling the input image while increasing channel capacity. We integrate XFABlock with CSP architecture in stages 2-4 to enhance multi-scale feature learning. The encoder neck processes multi-scale features through our SPGA module and employs GCFC3 modules for feature pyramid construction. The decoder remains largely unchanged from RT-DETR, using 300 learnable object queries and 4 decoder layers with 8 attention heads.

\subsection{Cross-Fuse Attention Block (XFABlock)}

The backbone requires enhanced feature extraction capabilities to capture the subtle patterns characteristic of pneumonia lesions. We propose XFAUnit, which employs a Cross Stage Partial (CSP) architecture combined with convolutional attention mechanisms to enhance multi-scale feature learning.

\begin{figure}[htbp]
\centering
\includegraphics[width=\columnwidth]{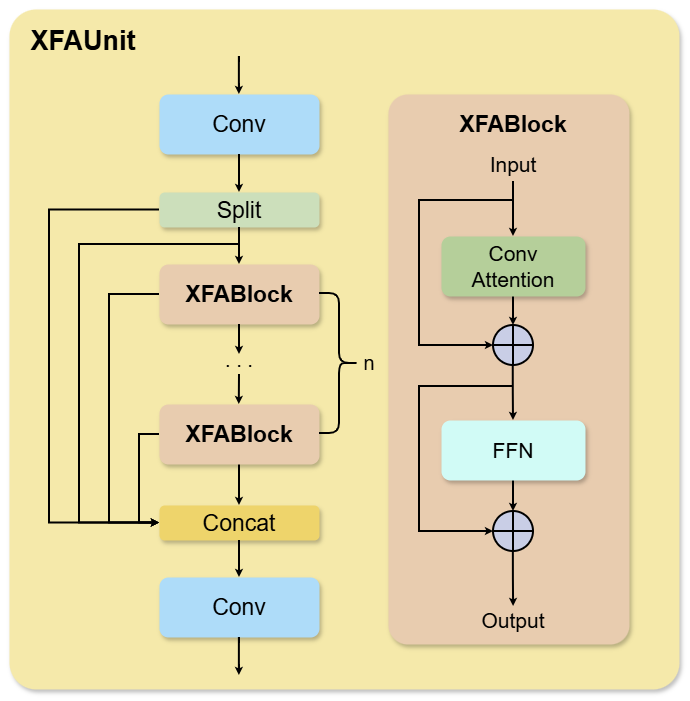}
\caption{Architecture of XFABlock. XFABlocks apply convolutional attention with residual and FFN branches inside a CSP-style structure to capture multi-scale context.}
\label{fig:xfablock}
\end{figure}

XFAUnit builds upon the CSP design principle, which splits features into two paths to improve gradient flow and reduce computational redundancy. Instead of using standard bottleneck structures, XFAUnit employs XFABlock modules that integrate convolutional attention for more expressive feature representations. The architecture is illustrated in Fig. \ref{fig:xfablock}.

Each XFABlock comprises a convolutional-attention transformation running in parallel with an identity residual shortcut, followed by a lightweight channel FFN. The convolutional attention employs multi-head depthwise separable \(3\times 3\) kernels to compute content-adaptive spatial weighting with low computational cost, while the residual pathway preserves information and stabilizes gradient flow. The subsequent FFN performs channel-wise mixing to refine the responses and enhance cross-channel interactions.

Formally, given an input feature map $X \in \mathbb{R}^{C \times H \times W}$, the XFABlock computes:

\begin{equation}
X' = X + \text{DropPath}(\text{ConvAttention}(X))
\end{equation}

\begin{equation}
X'' = X' + \text{DropPath}(\text{FFN}(X'))
\end{equation}

where ConvAttention denotes the convolutional attention operation and FFN is a two-layer perceptron with SiLU activation.

The complete XFAUnit module follows the CSP design pattern: it splits the input features along the channel dimension into two paths, processes one path through a sequence of XFABlocks while keeping the other as a shortcut, and then concatenates all intermediate features. This design enables the model to learn multi-level feature representations with varying receptive fields, which is crucial for detecting pneumonia lesions of different sizes and appearances.

\subsection{Split-Path Gated Attention (SPGA)}

The standard AIFI module in RT-DETR uses multi-head self-attention to enhance features within each scale. While effective, multi-head attention introduces significant computational overhead and may not be optimal for medical images where feature distributions differ from natural images. We propose SPGA that employs single-head self-attention with dynamic gating for focused feature aggregation.

\begin{figure*}[!t]
\centering
\includegraphics[width=0.9\textwidth]{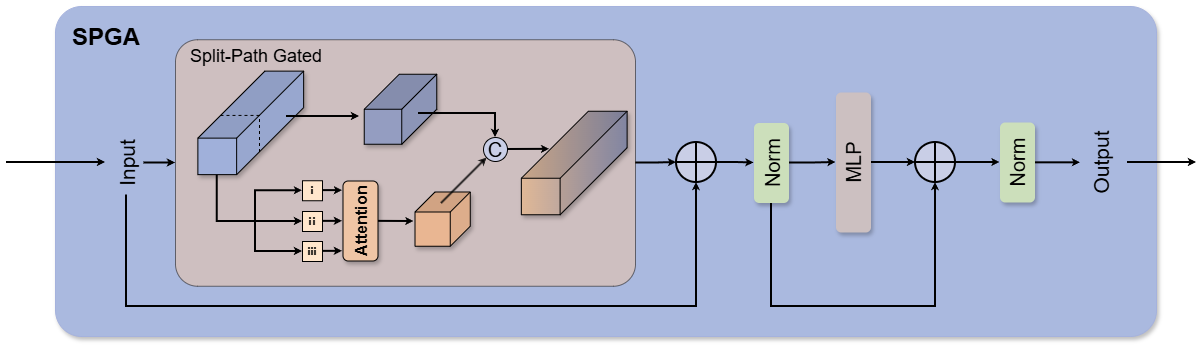}
\caption{SPGA module. A narrow branch applies single-head attention with dynamic sparsity while a wide bypass preserves high-frequency information before recombining. The gating network dynamically controls attention sparsity based on input content.}
\label{fig:spga}
\end{figure*}

SPGA, illustrated in Fig.~\ref{fig:spga}, first performs a channel-wise decomposition of the input feature map into a narrow attention branch and a wide bypass, i.e., $X=[X_1,X_2]$ with $X_1 \in \mathbb{R}^{C/4\times H\times W}$ and $X_2 \in \mathbb{R}^{3C/4\times H\times W}$. On the attention branch, group normalization precedes a single-head self-attention in which three projections denoted $Z^{(i)}$, $Z^{(ii)}$ and $Z^{(iii)}$ are produced by efficient $1\times1$ convolutions; $Z^{(i)}$ and $Z^{(ii)}$ are used to compute pairwise similarity, while $Z^{(iii)}$ provides the features to be aggregated. To adapt computational sparsity to image content, a gating network $G(\cdot)$ that consumes the full input $X$ predicts an instance-specific sparsity ratio determining the number of active attention links

\begin{equation}
k = \lfloor N \cdot \sigma(G(X)) \rfloor
\end{equation}

where $G$ is a lightweight convolutional module, $\sigma$ denotes the sigmoid function, and $N=H\times W$ is the number of spatial locations. The raw attention logits are then sparsified by preserving, for each query, only the top-$k$ entries and masking the remainder to negative infinity before softmax:

\begin{equation}
{Attention}_{ab} = \begin{cases}
\frac{\big(Z^{(i)}_a\big)^{\!T} Z^{(ii)}_b}{\sqrt{d}} & \text{if } b \in \text{TopK}\big(\big(Z^{(i)}_a\big)^{\!T} Z^{(ii)},\, k\big) \\
-\infty & \text{otherwise}
\end{cases}
\end{equation}

Here $\text{Attention}\in\mathbb{R}^{N\times N}$ denotes the masked attention logits computed from projections $Z^{(i)}$ and $Z^{(ii)}$ that relate position $a$ to position $b$; the aggregation projection $Z^{(iii)}$ does not appear in $\text{Attention}$ and is only used when forming the attended output. After row-wise normalization we obtain

\begin{equation}
\alpha_{ab} = \mathrm{softmax}_b\big(\text{Attention}_{ab}\big),\qquad
Y_a = \sum_{b=1}^{N} \alpha_{ab} \, Z^{(iii)}_b .
\end{equation}

For brevity we omit the batch index and the (optional) head index; in our single-head setting (SHSA) there is only one attention head. Spatial tokens are obtained by flattening the $H\times W$ feature map so that $N=H\times W$.

The attended responses computed from $X_1$ are finally concatenated with the bypass tensor $X_2$ and projected through a $1\times1$ convolution with SiLU activation, yielding the SPGA output that combines global contextual cues with preserved high-frequency details.

Taken together, this design improves representation quality and concentrates computation on informative interactions. While the single-head formulation is lighter than conventional multi-head attention, our ablations against RT-DETR's AIFI show that SPGA introduces mild overhead in exchange for accuracy gains; paired with GCFC3 the overall latency decreases. The gating network modulates sparsity in a data-adaptive way so computation concentrates on informative interactions; and the split-path topology preserves high-frequency details along the bypass while the attention branch aggregates long-range context, yielding features that are simultaneously sharp locally and coherent globally.

\subsection{Generalized Convolution Fusion RepC3 (GCFC3)}

Feature pyramid networks are essential for multi-scale object detection, but standard designs may not fully exploit multi-path feature fusion. We propose GCFC3 that enhances the RepC3 block through a multi-branch convolution fusion architecture with structural re-parameterization capability \cite{WangBochkovskiy2023}.

\begin{figure}[htbp]
\centering
\includegraphics[width=\columnwidth]{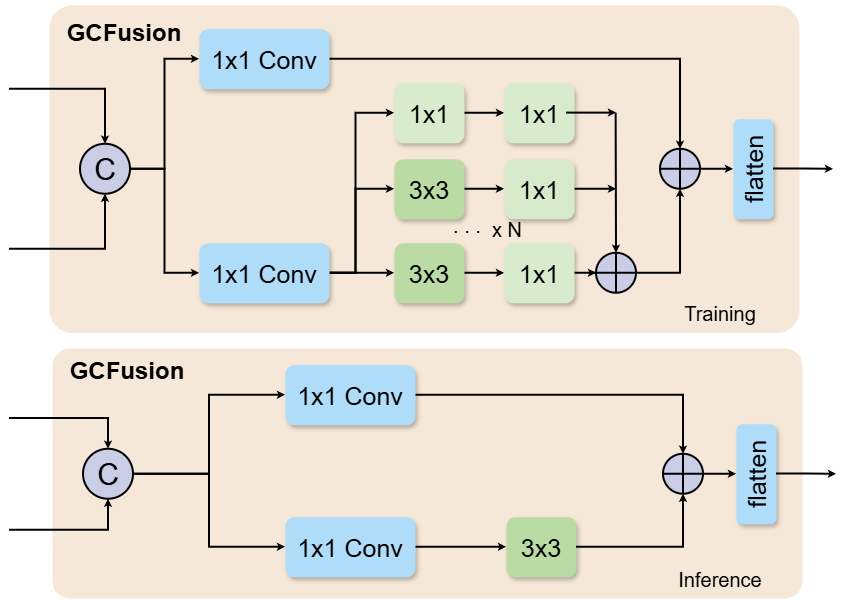}
\caption{GCFC3 module architecture. During training, multiple parallel convolution paths with diverse kernel configurations capture multi-scale features. During inference, these paths are structurally re-parameterized into a simplified form for computational efficiency.}
\label{fig:gcfc3}
\end{figure}

As illustrated in Fig. \ref{fig:gcfc3}, GCFC3 follows a CSP-like structure where the input features are first split along the channel dimension into a bypass branch $X_r$ and a transform branch $X_c$. One branch directly bypasses to preserve original information, while the other branch undergoes convolution transformations. During training, this transformation branch employs multiple parallel convolution paths with varying kernel configurations: 1$\times$1 convolutions for channel-wise transformations and cascaded 3$\times$3-1$\times$1 convolutions for spatial feature extraction. The outputs of these parallel paths within the transform branch are summed element-wise before activation, enabling the network to simultaneously capture multi-scale features through different receptive fields; at the end of the block, the transformed branch is concatenated with the bypass branch along the channel dimension and lightly fused.

Let the transform-branch aggregation be $T(X_c)=\sum_{i=1}^{M} P_i(X_c)$. The block output is then obtained by concatenating the transform and bypass branches followed by a pointwise fusion:

\begin{equation}
Y = \sigma(\, \Phi([\, T(X_c)\,;\, X_r \, ]))
\end{equation}

where $[\cdot\,;\,\cdot]$ denotes channel-wise concatenation, $\Phi(\cdot)$ is a $1\times1$ convolution used to mix channels and adjust dimensionality, $X_c$ and $X_r$ are the transform and bypass branches respectively, and $\sigma$ is the SiLU activation function.

A key advantage of GCFC3 lies in its structural re-parameterization strategy. During inference, the multiple parallel paths are algebraically fused into a simplified architecture. The batch normalization layers are absorbed into convolution weights, and kernels from different paths are merged through padding and summation:

\begin{equation}
W_{fused} = \sum_{i=1}^{M} \text{Transform}(W_i, BN_i)
\end{equation}

where Transform represents the re-parameterization operation. As shown in Fig. \ref{fig:gcfc3}, this results in a streamlined architecture with reduced computational complexity while preserving the representation capacity learned during training, achieving a balance between multi-scale feature learning and deployment efficiency.

\section{Experiments and Results}

\subsection{Experimental Setup}

We evaluate our method on the RSNA Pneumonia Detection dataset, which contains chest X-ray images with bounding box annotations for pneumonia regions. We follow the standard data split for training, validation, and testing.

Our implementation is based on PyTorch framework under Windows 11 system. The hardware configuration includes an NVIDIA GeForce RTX 4090 GPU with CUDA 12.1, and Python 3.10 deployment environment. Following the RT-DETR-l configuration, input images are resized to 1024x1024 pixels. The model is trained for 72 epochs using the AdamW optimizer with an initial learning rate of 1e-4 and batch size of 16. The loss function combines focal loss for classification \cite{Lin2017} and GIoU loss for bounding box regression.

We evaluate detection performance using standard metrics: mAP@0.5, mAP@[0.5:0.95], Recall@0.5, and Precision@0.5. Inference speed is measured in FPS, and we report model parameters (M) and FLOPs (G) at 1024$\times$1024 resolution.

\begin{table*}[t]
\centering
\caption{Comparison with other Object Detection Methods on RSNA Pneumonia Detection Dataset.}
\label{tab:comparison}
\resizebox{\textwidth}{!}{
\begin{tabular}{l|cccc|ccc|c|ccc}
\toprule
\textbf{Model} & \textbf{mAP@0.5} & \textbf{mAP@[0.5:0.95]} & \textbf{Recall@0.5} & \textbf{Precision@0.5} & \textbf{Params(M)} & \textbf{FLOPs(G)} & \textbf{Latency(ms)} & \textbf{FPS} & \textbf{$\Delta$mAP@0.5} & \textbf{$\Delta$mAP@[0.5:0.95]} & \textbf{$\Delta$Latency(ms)} \\
\midrule
Faster R-CNN R50-FPN & 73.5 & 44.5 & 77.5 & 72.5 & 42.0 & 190.0 & 32.8 & 30.5 & -5.0 & -3.6 & +11.5 \\
RetinaNet R50-FPN & 75.2 & 45.3 & 78.4 & 74.0 & 38.5 & 160.0 & 24.5 & 40.8 & -3.3 & -2.8 & +3.2 \\
YOLOv5-l & 76.0 & 46.0 & 79.0 & 76.0 & 46.5 & 155.0 & 20.3 & 49.3 & -2.5 & -2.1 & -1.0 \\
YOLOv7 & 76.8 & 46.3 & 79.4 & 76.5 & 36.9 & 152.0 & 19.8 & 50.5 & -1.7 & -1.8 & -1.5 \\
YOLOv8-l & 77.0 & 46.8 & 79.8 & 77.0 & 43.8 & 158.0 & 19.2 & 52.1 & -1.5 & -1.3 & -2.1 \\
YOLOv10-l & 77.4 & 47.2 & 80.0 & 77.5 & 45.2 & 150.0 & 18.5 & 54.1 & -1.1 & -0.9 & -2.8 \\
RTMDet-l & 77.2 & 46.6 & 79.6 & 76.8 & 52.0 & 162.0 & 20.0 & 50.0 & -1.3 & -1.5 & -1.3 \\
Deformable DETR R50 & 77.9 & 48.0 & 80.0 & 76.2 & 41.0 & 175.0 & 24.6 & 40.7 & -0.6 & -0.1 & +3.3 \\
DINO R50 & 79.0 & 49.0 & 81.0 & 79.0 & 47.5 & 178.0 & 23.2 & 43.1 & +0.5 & +0.9 & +1.9 \\
RT-DETR-l & 78.5 & 48.1 & 80.2 & 76.6 & 46.2 & 170.0 & 21.3 & 46.9 & +0.0 & +0.0 & +0.0 \\
\midrule
\textbf{CGF-DETR (Ours)} & \textbf{82.2} & \textbf{50.4} & \textbf{83.3} & \textbf{80.2} & 49.8 & 170.0 & 20.8 & 48.1 & \textbf{+3.7} & \textbf{+2.3} & \textbf{-0.5} \\
\bottomrule
\end{tabular}
}
\end{table*}

\begin{table*}[t]
	\centering
	\caption{Modular ablation experiments on RSNA Pneumonia Detection Dataset.}
	\label{tab:ablation}
	\resizebox{\textwidth}{!}{
		\begin{tabular}{cccc|cccc|ccc|c|ccc}
			\toprule
			\multicolumn{4}{c|}{\textbf{Components}} & \multicolumn{4}{c|}{\textbf{Performance Metrics}} & \multicolumn{3}{c|}{\textbf{Model Complexity}} & \textbf{Speed} & \multicolumn{3}{c}{\textbf{Improvements}} \\
			\cmidrule(lr){1-4} \cmidrule(lr){5-8} \cmidrule(lr){9-11} \cmidrule(lr){12-12} \cmidrule(lr){13-15}
			\textbf{XFA} & \textbf{SPGA} & \textbf{GCFC3} & & \textbf{mAP@0.5} & \textbf{mAP@[0.5:0.95]} & \textbf{Recall@0.5} & \textbf{Precision@0.5} & \textbf{Params(M)} & \textbf{FLOPs(G)} & \textbf{Latency(ms)} & \textbf{FPS} & \textbf{$\Delta$mAP@0.5} & \textbf{$\Delta$mAP@[0.5:0.95]} & \textbf{$\Delta$Latency} \\
			\midrule
			& & & Baseline & 78.5 & 48.1 & 80.2 & 76.6 & 46.2 & 170.0 & 21.3 & 46.9 & +0.0 & +0.0 & +0.0 \\
			\checkmark & & & +XFA & 80.1 & 49.0 & 81.2 & 78.4 & 48.6 & 176.8 & 22.5 & 44.4 & +1.6 & +0.9 & +1.2 \\
			& \checkmark & & +SPGA & 79.6 & 48.8 & 81.4 & 77.4 & 47.1 & 173.4 & 22.1 & 45.2 & +1.1 & +0.7 & +0.8 \\
			& & \checkmark & +GCFC3 & 79.2 & 48.6 & 80.8 & 77.3 & 46.5 & 159.8 & 19.0 & 52.6 & +0.7 & +0.5 & -2.3 \\
			\checkmark & \checkmark & & +XFA+SPGA & 81.4 & 49.8 & 82.4 & 79.2 & 49.5 & 180.2 & 23.4 & 42.7 & +2.9 & +1.7 & +2.1 \\
			\checkmark & & \checkmark & +XFA+GCFC3 & 80.9 & 49.6 & 81.8 & 79.1 & 48.9 & 166.6 & 20.1 & 49.8 & +2.4 & +1.5 & -1.2 \\
			& \checkmark & \checkmark & +SPGA+GCFC3 & 80.4 & 49.4 & 82.0 & 78.1 & 47.4 & 163.2 & 19.7 & 50.8 & +1.9 & +1.3 & -1.6 \\
			\checkmark & \checkmark & \checkmark & \textbf{CGF-DETR} & \textbf{82.2} & \textbf{50.4} & \textbf{83.3} & \textbf{80.2} & 49.8 & 170.0 & 20.8 & 48.1 & \textbf{+3.7} & \textbf{+2.3} & \textbf{-0.5} \\
			\bottomrule
		\end{tabular}
	}
\end{table*}

\subsection{Comparative Experiments}

Table \ref{tab:comparison} presents a comparison of CGF-DETR with well-established object detection methods on the RSNA Pneumonia Detection dataset. Our method achieves 82.2\% mAP@0.5 and 50.4\% mAP@[0.5:0.95], outperforming both CNN-based detectors and transformer-based detectors.

Compared to the baseline RT-DETR-l, CGF-DETR improves mAP@0.5 by 3.7\% and mAP@[0.5:0.95] by 2.3\% while actually reducing inference latency by 0.5ms. This demonstrates that our proposed modules not only enhance accuracy but also maintain the real-time performance critical for practical deployment.

Against recent YOLO variants, CGF-DETR achieves substantially higher accuracy with competitive speed. Compared to transformer-based methods like DINO, our approach achieves 3.2\% higher mAP@0.5 while being significantly faster.

The high Recall@0.5 of 83.3\% and Precision@0.5 of 80.2\% demonstrate that CGF-DETR effectively balances false negatives and false positives, which is critical in medical applications where both missed detections and false alarms have clinical consequences.

\subsection{Ablation Experiments}

To analyze the contribution of each proposed component, we conduct comprehensive ablation studies by progressively adding modules to the baseline RT-DETR-l architecture. Table \ref{tab:ablation} presents the results.

Individual module analysis shows XFABlock contributes the largest gain, while SPGA adds +1.1\% with a small compute increase. Notably, GCFC3 improves accuracy by +0.7\% while reducing latency by 2.3ms through structural re-parameterization.

Two-module combinations demonstrate synergistic effects: XFA+SPGA achieves +2.9\% mAP@0.5, XFA+GCFC3 reaches +2.4\% with improved speed, and SPGA+GCFC3 attains +1.9\% with the lowest latency among dual configurations.

The complete CGF-DETR achieves 82.2\% mAP@0.5 with 20.8ms latency, lower than XFA alone or XFA+SPGA. The combined gain slightly exceeds the sum of individual contributions, confirming complementary benefits across components where GCFC3's efficiency offsets overhead from the attention modules.

\subsection{Visualization Analysis}

\begin{figure}[htbp]
\centering
\includegraphics[width=\columnwidth]{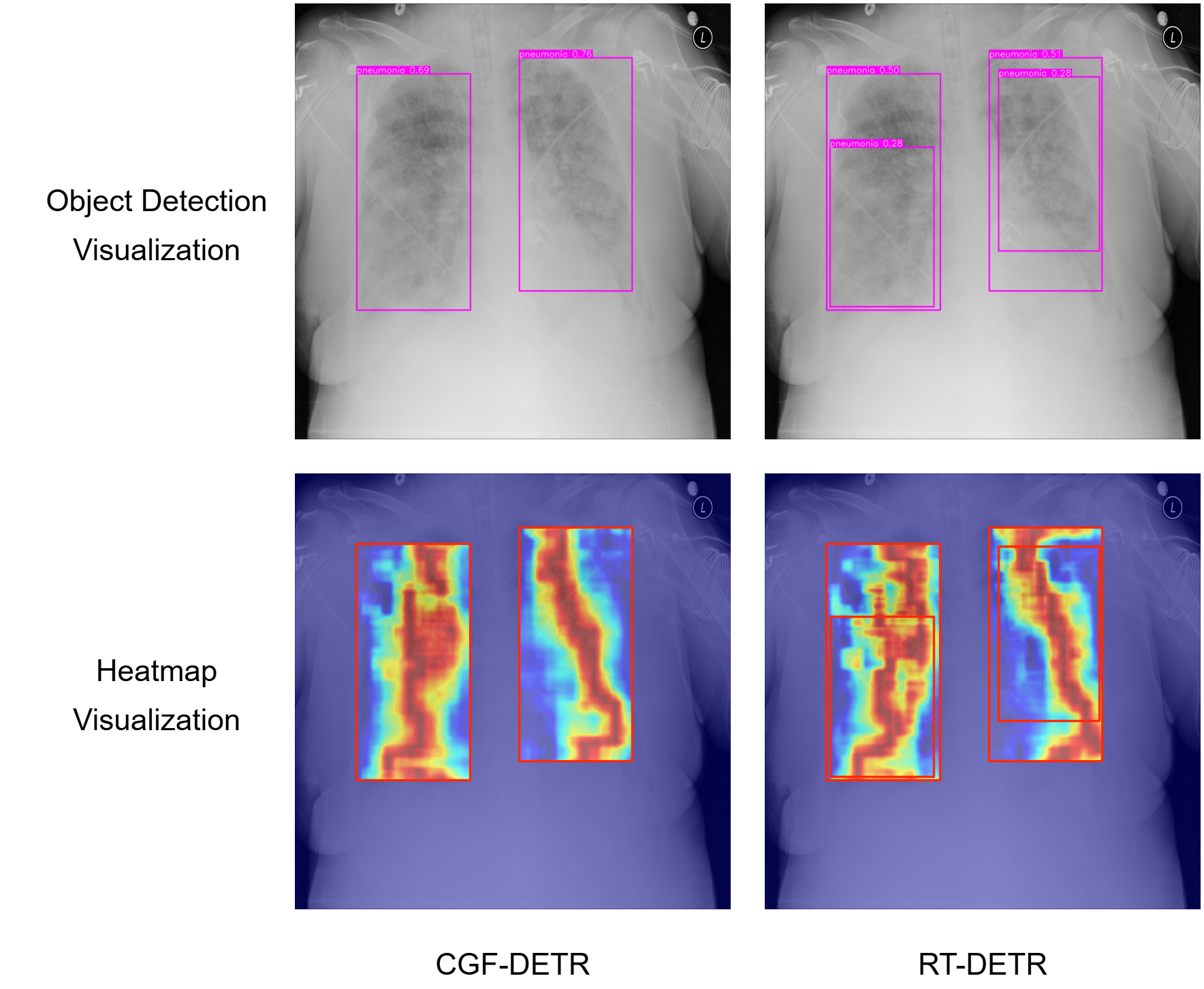}
\caption{Visualization comparison between CGF-DETR and RT-DETR.}
\label{fig:visualization}
\end{figure}

Fig. \ref{fig:visualization} presents a qualitative comparison between RT-DETR-l and our CGF-DETR, showing both detection results and attention heatmap visualizations.

As shown in Fig. \ref{fig:visualization}, CGF-DETR produces more accurate detection results with higher confidence scores and fewer redundant predictions. While both models detect the pneumonia regions, our model generates 2 precise detections with high confidence, whereas RT-DETR produces 4 overlapping detections with lower confidence scores, indicating less reliable predictions. The attention heatmap visualization reveals that CGF-DETR focuses more precisely on pneumonia-affected regions with concentrated attention distribution. In contrast, RT-DETR shows more diffuse attention patterns spread across larger areas, which contributes to the redundant and less confident detections.

\section{Conclusion}

This paper presents CGF-DETR, an enhanced real-time transformer for pneumonia detection in chest X-rays. We introduce three synergistic modules: XFABlock integrates convolutional attention with CSP architecture for improved multi-scale feature extraction, SPGA employs gated single-head attention for focused feature fusion, and GCFC3 enhances multi-scale representation through structural re-parameterization. Experiments on the RSNA dataset show CGF-DETR achieves 82.2\% mAP@0.5 and 50.4\% mAP@[0.5:0.95] while maintaining real-time performance at 48.1 FPS. Ablation studies confirm each module's contribution; although SPGA alone introduces mild overhead, the complete model demonstrates effective synergy where GCFC3's efficiency offsets attention-related cost, resulting in lower end-to-end latency than the baseline. These domain-specific enhancements for capturing subtle pneumonia patterns may provide insights for broader medical imaging applications.

\section*{Acknowledgment}

We thank the RSNA and the Kaggle community for providing the pneumonia detection dataset.

\bibliographystyle{IEEEtran}
\bibliography{references}

\end{document}